\documentclass[letterpaper]{article} 
\usepackage{aaai2026}  
\usepackage{times}  
\usepackage{helvet}  
\usepackage{courier}  
\usepackage[hyphens]{url}  
\usepackage{graphicx} 
\usepackage{natbib}  
\usepackage{caption} 
\usepackage{algorithm}
\usepackage{algorithmic}
\usepackage{amsmath}
\usepackage{url}
\usepackage{colortbl}
\usepackage{amssymb}

\usepackage{newfloat}
\usepackage{listings}
\DeclareCaptionStyle{ruled}{labelfont=normalfont,labelsep=colon,strut=off} 
\floatstyle{ruled}
\newfloat{listing}{tb}{lst}{}
\floatname{listing}{Listing}
\title{OR-R1: Automating Modeling and Solving of Operations Research Optimization Problem via Test-Time Reinforcement Learning}
\author {
    Zezhen Ding\textsuperscript{\rm 1}, 
    Zhen Tan\textsuperscript{\rm 2}, 
    Jiheng Zhang\textsuperscript{\rm 1}\correspondingauthor, 
    Tianlong Chen\textsuperscript{\rm 3}\correspondingauthor
}
\affiliations {
    \textsuperscript{\rm 1}The Hong Kong University of Science and Technology\\
    \textsuperscript{\rm 2}Arizona State University\\
    \textsuperscript{\rm 3}University of North Carolina at Chapel Hill\\
    zdingah@connect.ust.hk, ztan36@asu.edu, jiheng@ust.hk, tianlong@cs.unc.edu
}

\usepackage{bibentry}

\begin{document}

\maketitle

\begin{abstract}
Optimization modeling and solving are fundamental to the application of Operations Research (OR) in real-world decision making, yet the process of translating natural language problem descriptions into formal models and solver code remains highly expertise intensive. While recent advances in large language models (LLMs) have opened new opportunities for automation, the generalization ability and data efficiency of existing LLM-based methods are still limited, asmost require vast amounts of annotated or synthetic data, resulting in high costs and scalability barriers. In this work, we present OR-R1, a data-efficient training framework for automated optimization modeling and solving. OR-R1 first employs supervised fine-tuning (SFT) to help the model acquire the essential reasoning patterns for problem formulation and code generation from limited labeled data. In addition, it improves the capability and consistency through Test-Time Group Relative Policy Optimization (TGRPO). This two-stage design enables OR-R1 to leverage both scarce labeled and abundant unlabeled data for effective learning. Experiments show that OR-R1 achieves state-of-the-art performance with an average solving accuracy of $67.7\%$, using only $1/10$ the synthetic data required by prior methods such as ORLM, exceeding ORLM’s solving accuracy by up to $4.2\%$. Remarkably, OR-R1 outperforms ORLM by over $2.4\%$ with just $100$ synthetic samples. Furthermore, TGRPO contributes an additional $3.1\%–6.4\%$ improvement in accuracy, significantly narrowing the gap between single-attempt (Pass@1) and multi-attempt (Pass@8) performance  from $13\%$ to $7\%$. Extensive evaluations across diverse real-world benchmarks demonstrate that OR-R1 provides a robust, scalable, and cost-effective solution for automated OR optimization problem modeling and solving, lowering the expertise and data barriers for industrial OR applications.
\end{abstract}

\begin{links}
    \link{Code}{https://github.com/SCUTE-ZZ/OR-R1}
\end{links}

\begin{figure}[t]
\centering
\includegraphics[width=0.45\textwidth]{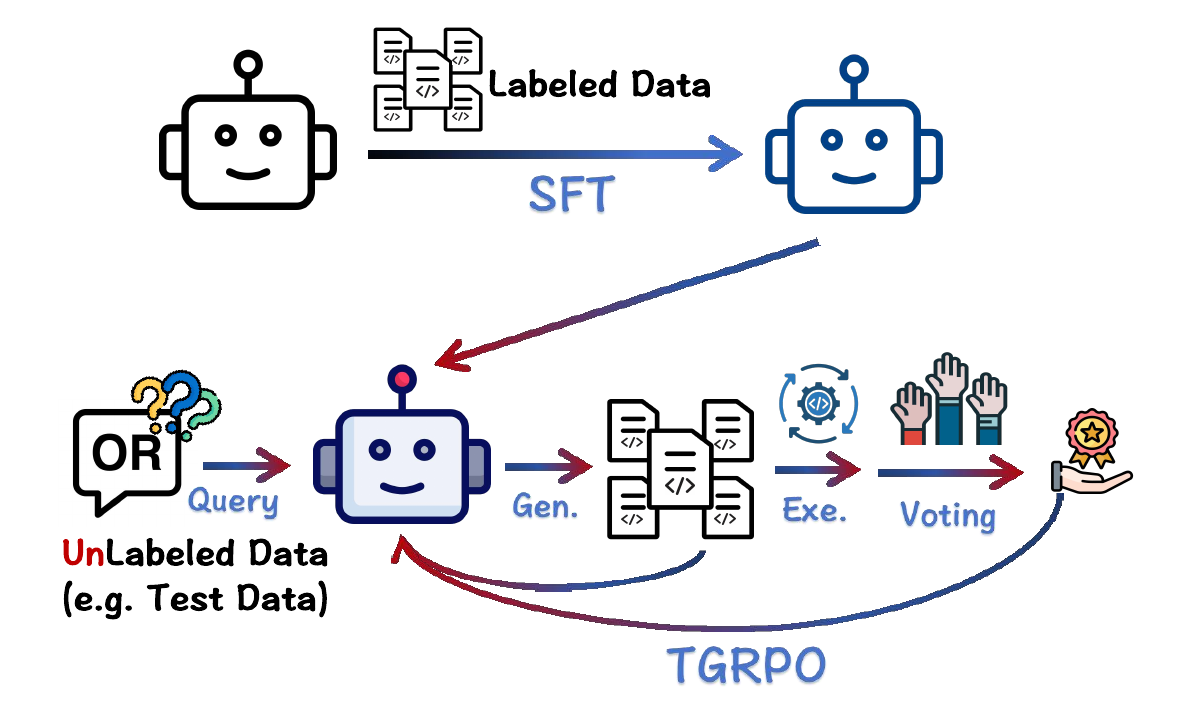} 
\caption{Overview of OR-R1.}
\label{fig0}
\end{figure}

\section{Introduction}
The field of artificial intelligence has witnessed remarkable advancements, with Large Language Models (LLMs) emerging as powerful tools across diverse domains~\cite{zhao2025surveylargelanguagemodels,he2024olympiadbench,jiang2024surveylargelanguagemodels}. A particularly promising, yet challenging, application lies in automating the modeling and solving of optimization problems~\cite{ramamonjison2023nl4opt}. These problems are central to numerous scientific and industrial applications, including logistics~\cite{lee2015impact, harrison2019logistics}, resource allocation~\cite{bretthauer1995nonlinear}, and scheduling~\cite{brucker1999resource, long2020dynamic}, where even minor improvements can yield significant real-world benefits. Traditionally, formulating these problems into precise mathematical models and subsequently generating executable solver code has demanded specialized human expertise, often a time-consuming and error-prone process~\cite{huang2025orlm, jiang2024llmopt}. LLMs have emerged as a promising tool to automate this process, reducing the expertise barrier for solving optimization problems.

Recent research has explored the integration of LLMs into the optimization pipeline, broadly falling into two main categories: prompt-based approaches and learning-based approaches. Prompt-based approaches typically leverage LLMs' in-context learning capabilities through carefully designed prompts, few-shot examples, or chain-of-thought methods to generate optimization models or code directly. Notable works in this category include the Chain-of-Experts \cite{xiao2023chain}, Optimus \cite{ahmaditeshnizi2024optimus}, MAMO \cite{huang2024mamo}, ORQA \cite{mostajabdaveh2025evaluating}, OR-LLM-Agent \cite{zhang2025or}, and OptimAI \cite{thind2025optimai}. These methods often utilize general-purpose LLMs without extensive domain-specific fine-tuning, relying on their pre-trained knowledge and sophisticated prompting strategies. In contrast, learning-based approaches involve fine-tuning or training LLMs on domain-specific datasets to enhance their understanding and generation capabilities for optimization problems. Key contributions in this area include ORLM \cite{huang2025orlm}, LLMOPT \cite{jiang2024llmopt}, and OptiBench \cite{yang2024optibench}. These works aim to build more specialized and performant models through data-driven learning.

Learning-based approaches hold great promise for automated optimization; for instance, ORLM, using a Llama3-8B model, has even surpassed GPT-4o \cite{hurst2024gpt} in some tests. Despite this, a major hurdle remains: the heavy reliance on vast amounts of domain-specific data. While synthetic data can be mass-produced, it often lacks the real-world rigor and diversity of human-made examples. Conversely, manual annotation of high-quality data is extremely costly, leading to a shortage of large, useful datasets and limiting adoption. Our main goal is to significantly cut down on the amount of labeled data needed. Moreover, research, including insights from ORLM, shows another key challenge: LLMs can find optimal solutions, but their single-attempt outputs often lack consistency. This means they're more likely to get the right answer if they generate multiple times. This inconsistency highlights a clear need to improve the reliability of individual outputs.

To address these critical issues, the large data demands and inconsistent outputs, we introduce OR-R1. This innovative learning-based framework is specifically designed to boost LLM performance in Operations Research by requiring far less data and delivering greater consistency. First, OR-R1 uses Supervised Fine-Tuning (SFT) to teach the core reasoning of the model for optimization modeling and code generation with limited labeled data. Then, it integrates Test-Time Group Relative Policy Optimization (TGRPO). TGRPO works by having the LLM predict labels for unlabeled data, using a voting system to create high-quality pseudo-labels. These pseudo-labels then serve as a reward function for reinforcement learning. This two-part strategy not only drastically cuts the need for expensive labeled training data but also substantially improves the consistency of the model's predictions, making single-attempt generations much more reliable.

Our main contributions and findings are:
\begin{itemize}

\item We introduce \textbf{OR-R1}, a novel framework that, for the first time, integrates SFT and \textbf{TGRPO} for automated operations research problems modeling and solving.

\item We design a multi-faceted reward system specifically tailored for optimization problem scenarios, comprising \textbf{Format Reward} for structural correctness, \textbf{Valid-Code Reward} for executability, \textbf{Majority Voting Reward} for numerical accuracy, enabling robust model learning.

\item OR-R1 surpasses previous state-of-the-art methods using only 1/10 synthetic data of ORLM, and attains an average solving accuracy of \textbf{67.7\%} across diverse public benchmarks, while significantly narrowing the gap between single-attempt and multi-attempt accuracy.
    
\end{itemize}


\begin{figure*}[t]
\centering
\includegraphics[width=0.95\textwidth]{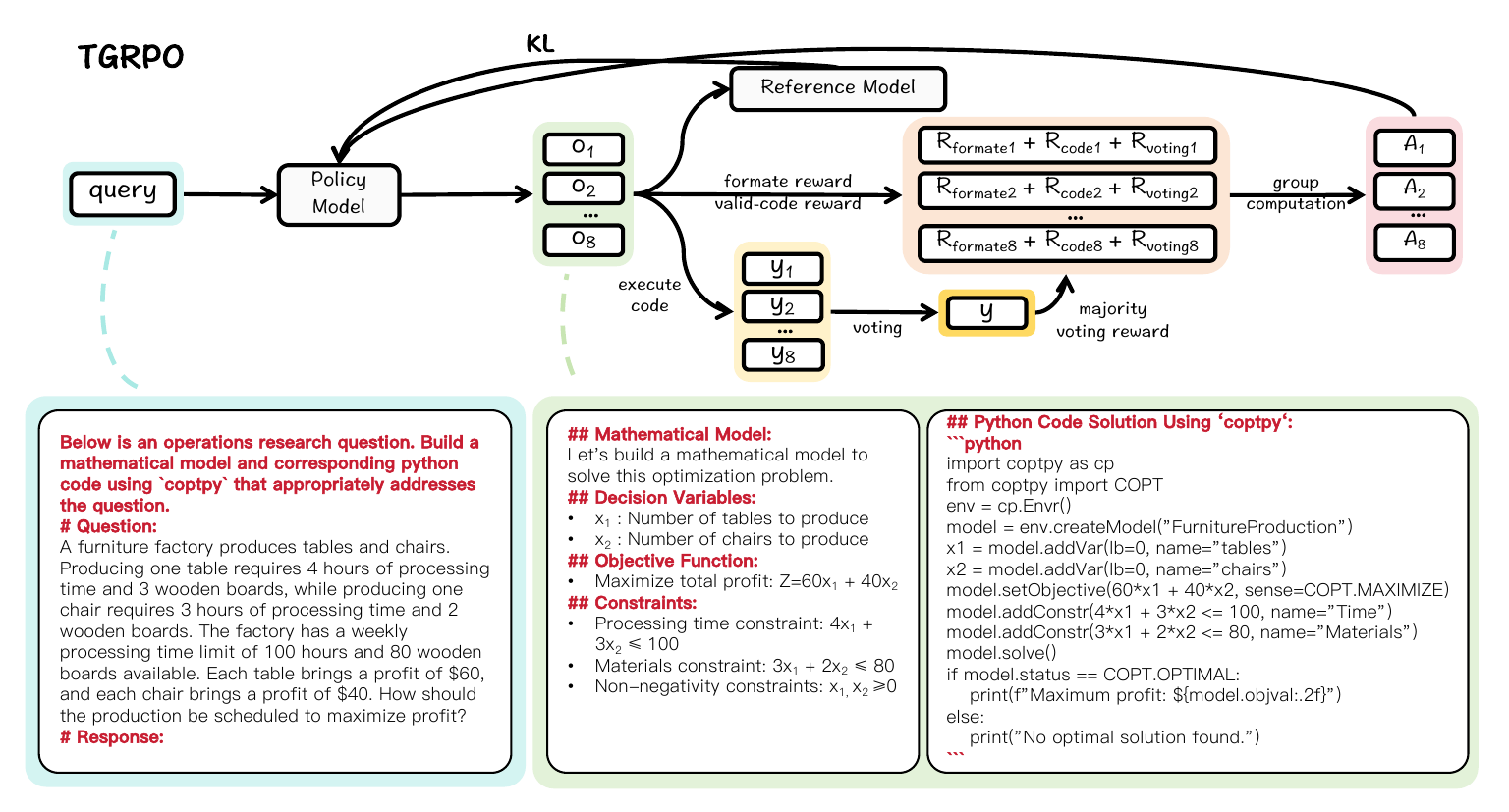} 
\caption{The figure illustrates the training process of TGRPO. The blue section shows an example of an operations research (OR) problem. The green section presents sample outputs, including the mathematical model and corresponding code. The light yellow section displays the code execution results, while the yellow section shows the majority voting of execution results. The orange section represents the reward function, and the red section indicates the advantage function.}
\label{fig1}
\end{figure*}

\section{Related Work}

\noindent\textbf{Automated OR Problem Modeling and Solving.}
Large Language Models (LLMs) have demonstrated impressive capabilities in formal reasoning and code generation, advancing from solving math word problems \cite{cobbe2021training,hendrycks2021measuring,ahn2024largelanguagemodelsmathematical} to generating competition-level code \cite{li2022competition,codealpaca,bairi2024codeplan}. Meanwhile, methods such as few-shot learning \cite{brown2020language}, Chain-of-Thought prompting \cite{wei2022chain}, Tree-of-Thoughts \cite{yao2023tree}, and Graph-of-Thoughts \cite{besta2024graph} have significantly enhanced the reasoning abilities of Large Language Models (LLMs). Building upon these foundational advancements in general math and coding, a specialized and highly impactful application of LLMs has emerged in the domain of Operations Research (OR). Specifically, LLMs are now being used for automated optimization modeling. This involves translating natural language descriptions of real-world problems into precise mathematical optimization formulations (e.g., MILP, LP) or directly into executable solver code. Research in this area can be broadly categorized into prompt-based and learning-based approaches. Prompt-based methods leverage large pretrained LLMs with sophisticated prompting strategies to generate models. Notable examples include the NL4Opt competition \cite{ramamonjison2023nl4opt}, which showcased LLMs' potential in understanding problem descriptions, and frameworks such as Optimus \cite{ahmaditeshnizi2024optimus}, MAMO \cite{huang2024mamo}, ORQA \cite{mostajabdaveh2025evaluating}, OR-LLM-Agent \cite{zhang2025or}, OptMATH \cite{lu2025optmath}, and OptimAI \cite{thind2025optimai}, which explore various prompting strategies, agent-based systems, and benchmarks for OR problem solving. In contrast, learning-based approaches involve fine-tuning LLMs on domain-specific datasets to achieve deeper understanding and generation capabilities. Key contributions include ORLM \cite{huang2025orlm}, LLMOPT \cite{jiang2024llmopt}, and OptiBench \cite{yang2024optibench}, which focuses on creating specialized models or benchmarks for this task. However, these methods typically rely on large amounts of synthetic data or costly human-annotated data for training. Synthetic data often suffers from accuracy issues, as highlighted by ORLM, where their synthetic data had an accuracy of only around 70\%. In contrast, acquiring high-quality human-annotated data is expensive and time consuming. 

\noindent\textbf{RL for LLM Alignment and Reasoning.}
Reinforcement Learning (RL) has become a cornerstone for aligning LLMs with human preferences and improving their reasoning capabilities. Initial works like RLHF \cite{ouyang2022training, christiano2017deep} and Direct Preference Optimization (DPO) \cite{rafailov2023direct} have shown the effectiveness of fine-tuning LLMs with human feedback. More advanced RL techniques, often based on Proximal Policy Optimization (PPO) \cite{schulman2017proximal}, are used to refine model behaviors for complex tasks. DeepSeekMath \cite{shao2024deepseekmath}, which introduced Group Relative Policy Optimization (GRPO), demonstrated how RL can push the limits of mathematical reasoning by learning from comparisons between multiple generated solutions. Subsequent work, DeepSeek-R1 \cite{guo2025deepseek}, further explores incentivizing reasoning capabilities via RL. Other research has focused on scaling RL systems for LLMs \cite{yu2025dapo, sheng2024hybridflow, zhang2024framework}, self-correction mechanisms \cite{kumar2024training, qu2024recursive} or formal provers \cite{xin2024deepseek} to guide RL training. 

\noindent\textbf{Test-Time Adaptation and Confidence in LLMs.}
Ensuring the reliability and accuracy of LLM outputs, particularly at inference time, is crucial. In complex problem-solving tasks, such as mathematical reasoning and code generation, it's commonly observed that approaches employing multiple votings over generated candidate solutions, followed by selection of the most consistent or accurate, achieve significantly better results than single-attempt methods (Pass@1) \cite{chen2023universalselfconsistencylargelanguage, huang2024enhancinglargelanguagemodels}. This phenomenon highlights that improving the consistency of generated outputs can lead to a substantial enhancement in Pass@1 performance. To address this, Test-Time Adaptation (TTA) methods aim to adapt models to new data distributions or improve performance. Examples include TENT \cite{wang2020tent}, which minimizes entropy during inference. Building on this, Test-Time Reinforcement Learning \cite{zuo2025ttrl,prabhudesai2025maximizing,yu2025dapo, sheng2024hybridflow, zhang2024framework, kumar2024training, qu2024recursive, xin2024deepseek} extends TTA by applying reinforcement learning principles to adapt models at test time.  By employing strategies like majority voting among candidate outputs and learning from consistency signals, these methods enable models to adapt on-the-fly to distribution shifts or challenging instances without extensive retraining. Despite these advances, such RL and adaptation techniques have not yet been systematically explored for solving operations research (OR) problems. In this work, we conduct the first application of these methods to automated OR problem solving.

\section{Method}

\subsection{Overview} 
The overall workflow of the OR-R1 training procedure is illustrated in Figure~\ref{fig0}. Our method utilizes the Qwen3-8B model as the base. First, we perform supervised fine-tuning (SFT) using a small, randomly selected subset of data from the 3000 IndustryOR dataset, an open-source resource from ORLM. Second, the model undergoes further training with TGRPO on unlabeled test set data. TGRPO training is guided by a composite reward function specifically designed for OR problem modeling and solving. This reward integrates several components derived from the generated outputs: (1) Format Reward, which measures adherence to the required structural format; (2) Valid-Code Reward, which checks the syntactic correctness and executability of the generated code; and (3) Majority Voting Reward, which reflects the consensus among multiple candidate solutions through majority voting. By combining these criteria, the reward function incentivizes the model to generate well-structured, functional, and consistent solutions for OR tasks.

\subsection{Supervised Fine-tuning (SFT) Phase}
The objective of the SFT phase is to maximize the likelihood of generating the correct output given the input. This is achieved by minimizing the negative logarithmic likelihood loss, a standard approach in supervised learning. The objective function for SFT is formally defined as:
\begin{equation}
\mathcal{L}_{\text{SFT}}(\theta) = - \mathbb{E}_{(x,o) \sim \mathcal{D}_{\text{SFT}}} \left[ \sum_{t=1}^{|o|} \log P(o_t | x, o_{<t}; \theta) \right]
\end{equation}

where
\begin{itemize}
    \item $\mathcal{L}_{\text{SFT}}(\theta)$ represents the loss function for the model parameters $\theta$.
    \item $\mathbb{E}_{(x,o) \sim \mathcal{D}_{\text{SFT}}}$ denotes the expectation over the input-output pairs $(x,o)$ sampled from the supervised data set $\mathcal{D}_{\text{SFT}}$, where $x$ is the input prompt and $o = (o_1, o_2, \ldots, o_{|o|})$ is the target output sequence.
    \item $P(o_t | x, o_{<t}; \theta)$ is the probability of generating the $t$-th token $o_t$ given the input prompt $x$, all preceding tokens $y_{<t}$, and the model parameters $\theta$.
\end{itemize}


\subsection{Test-Time Group Relative Policy Optimization (TGRPO)}
The theoretical foundation of TGRPO is consistent with that of GRPO. The training process is illustrated in Figure \ref{fig1}. A key characteristic of TGRPO is its ability to forego a separate critic model, instead estimating the baseline from group scores, significantly reducing computational training resources. The objective of TGRPO is to maximize the objective of the policy, driving the model to produce higher-quality outputs based on reward signals. The objective function for TGRPO is formally defined as:

\begin{equation}
\begin{split}
& \mathcal{J}_{TGRPO}(\theta) = \mathbb{E}_{q \sim P(Q), \{o_i\}_{i=1}^G \sim \pi_{\theta_{old}}(O|q)} \Big[ \frac{1}{G} \sum_{i=1}^G \big( \\
& \qquad \min \big( \frac{\pi_\theta(o_i|q)}{\pi_{\theta_{old}}(o_i|q)} A_i, \text{clip} \big( \frac{\pi_\theta(o_i|q)}{\pi_{\theta_{old}}(o_i|q)}, 1-\epsilon, 1+\epsilon \big) A_i \big) \\
& \qquad - \beta \mathbb{D}_{KL}(\pi_\theta || \pi_{ref}) \big) \Big]
\end{split}
\end{equation}

\begin{equation}
\begin{split}
& \mathbb{D}_{KL}(\pi_\theta||\pi_{ref}) = \frac{\pi_{ref}(o_i|q)}{\pi_\theta(o_i|q)} - \log \frac{\pi_{ref}(o_i|q)}{\pi_\theta(o_i|q)} - 1 
\end{split}
\end{equation}

\begin{equation}
\begin{split}
A_i = \frac{R_i - \text{mean}(\{R_1, R_2, \cdots, R_G\})}{\text{std}(\{R_1, R_2, \cdots, R_G\})}
\end{split}
\end{equation}

where 
\begin{itemize}
    \item $\mathcal{J}_{\text{TGRPO}}(\theta)$: Represents the objective function for the policy model with parameters $\theta$.
    \item $\mathbb{E}_{q \sim P(Q), \{o_i\}_{i=1}^G \sim \pi_{\theta_{\text{old}}}(O|q)}$: Denotes the expectation over questions $q$ sampled from the distribution $P(Q)$ and groups of outputs $\{o_i\}_{i=1}^G$ sampled from the old policy $\pi_{\theta_{\text{old}}}$ given $q$.
    \item $G$: The number of generated outputs in a group.
    \item $\pi_\theta(o_i|q)$: The probability of generating output $o_i$ given question $q$ under the current policy $\pi_\theta$.
    \item $\pi_{\theta_{\text{old}}}(o_i|q)$: The probability of generating output $o_i$ given question $q$ under the old policy $\pi_{\theta_{\text{old}}}$.
    \item $\pi_{\text{ref}}$: A fixed reference policy, often the SFT model or an initial pre-trained model, used for the KL divergence regularization.
    \item $R_i$: The reward for the output $o_i$.
    \item $A_i$: The advantage for output $o_i$.
    \item $\epsilon$: A hyperparameter for the PPO-style clipping.
    \item $\beta$: A hyperparameter controlling the strength of the KL divergence penalty.
    \item $\mathbb{D}_{\text{KL}}(\pi_\theta || \pi_{\text{ref}})$: The KL divergence between the current policy $\pi_\theta$ and a reference policy $\pi_{\text{ref}}$.
\end{itemize}

\subsection{Reward Function}
Our training framework incorporates a composite reward function, combining several distinct reward components to guide the model's learning process:

\subsubsection{Format Reward.}
The Format Reward encourages the model to generate outputs that adhere to a predefined structural or syntactical format. Specifically, we check for the presence of six key fields: ‘\#\# Mathematical Model:’, ‘\#\# Decision Variables:’, ‘\#\# Objective Function:’, ‘\#\# Constraints:’, ‘\#\# Python Code Solution Using \texttt{`}coptpy\texttt{`}:’, and ‘\texttt{```}python’. The reward is calculated as the proportion of these fields successfully identified in the output:
\begin{equation}
R_{\text{format}}(o_i) = \frac{\text{Number of required fields found}}{6}
\end{equation}

\subsubsection{Valid-Code Reward.}
The Valid-Code Reward incentivizes the generation of executable or syntactically correct code, crucial for tasks involving code generation or problem-solving through programmatic means. This is a binary reward:
\begin{equation}
R_{\text{code}}(o_i) = \begin{cases} 1, & \text{if code can correctly call \texttt{`}coptpy\texttt{`}} \\ 0, & \text{otherwise} \end{cases}
\end{equation}

\subsubsection{Majority Voting Reward.}
The Majority Voting Reward is derived from the Test-Time Reinforcement Learning (TTRL) framework \cite{zuo2025ttrl}, where a consensus output is established through majority voting among multiple candidate generations. This estimated consensus output then serves as a proxy label to compute a rule-based reward. The reward function is defined as:
\begin{equation}
R_{\text{voting}}(y_{i},y) = \begin{cases} 1, & \text{if } y_{i} = y \\ 0, & \text{otherwise} \end{cases}
\end{equation}
where:
\begin{itemize}
    \item $y_{i}$ denotes the optimal solution by executing the code.
    \item $y$ denotes the majority-voted prediction or the consensus output. For majority voting, only results that produce normal values from code execution are considered, ignoring cases with ‘No Best Solution’ or ‘None’ results.
\end{itemize}

The final composite reward $R$ is the sum of these individual reward components:
\begin{equation}
R_i=R_{\text{format}}(o_i) + R_{\text{code}}(o_i) + R_{\text{voting}}(y_{i},y)
\end{equation}

\section{Experiments}

\begin{table*}[h!]
\renewcommand{\arraystretch}{1.5}
\centering
\resizebox{\textwidth}{!}
{
\begin{tabular}{lccccccccc}
\hline
\rowcolor{blue!10} 
\textbf{Model} & \textbf{NL4OPT} & \textbf{MAMO} & \textbf{MAMO} & \textbf{IndustryOR} & \textbf{NLP4LP} & \textbf{ComplexOR} & \textbf{OptiBench} &\textbf{ICML} & \textbf{AVG}\\
\rowcolor{blue!10} 
& & \textbf{EasyLP} & \textbf{ComplexLP} & & & & & \textbf{Competition} &\\
\hline
\multicolumn{10}{c}{\textit{Base Model Variants}} \\
\hline

Qwen3-8B SFT(3K) & 86.0±2.0 & 87.0±1.0 & 39.9±3.0 & \underline{33.0±1.0} & 82.9±0.5 & 40.7±6.4 & \underline{61.4±1.2} &
\underline{85.8±2.0} & 64.6±1.2\\

\rowcolor{gray!20} 
Qwen2.5-7B SFT(3K) & 83.0±1.9 & 85.6±0.7 & 37.3±1.2 & 32.7±0.6 & 80.0±0.7 & 40.7±3.2 & 57.0±1.6
& 79.2±1.8 & 61.9±1.2\\

Llama3-8B SFT(3K) & 80.3±3.6 & 81.7±2.1 & 32.2±2.6 & 24.7±3.1 & 78.5±1.8 & 37.0±8.5 & 54.4±2.2
& 77.1±1.3 & 58.2±1.7\\

\rowcolor{gray!20} 

Qwen3-8B SFT(100) & 81.8±1.7 & 84.4±3.3 & 31.9±3.1 & 29.3±4.2 & 78.2±0.9 & 35.2±3.2 & 56.2±2.0 
& 79.4±2.8 & 59.5±1.7\\

\hline
\multicolumn{10}{c}{\textit{Learning Base Model}} \\
\hline

LLMOPT(Qwen2.5-14B) & 80.3 & \textbf{89.5} & 44.1 & 29.0 & 73.4 & 35.3 & 53.8 & 75.3 & 60.1\\

\rowcolor{gray!20} 
ORLM(Llama3-8B) & 86.9 & 81.6 & 39.3 & 32.0 & 82.0 & \textbf{50.0} & 56.5  & 79.3 & 63.5\\

\hline
\multicolumn{10}{c}{\textit{Proposed Method}} \\
\hline

OR-R1 SFT(100)-TGRPO & \underline{88.0±0.7} & \underline{87.4±2.5} & \underline{45.7±8.5} & 30.3±3.1 & \underline{84.0±1.0} & \underline{46.3±8.5} & 61.2±0.7 & 84.1±1.5 & \underline{65.9±2.2}\\

\rowcolor{gray!20} 
\textbf{OR-R1 SFT(3K)-TGRPO} & \textbf{88.3±0.9} & 86.1±1.0 & \textbf{49.9±15.0} & \textbf{35.3±2.9} & \textbf{84.6±0.8} & \underline{46.3±3.2} & \textbf{62.9±1.0}
& \textbf{88.3±1.8} & \textbf{67.7±2.7}\\

\hline

\end{tabular}
}

\caption{Main evaluation results on eight operations research benchmarks. Solution accuracy (\%) is reported for each method, with the overall average (AVG) in the last column. Bold indicates the best result. Values with ‘±’ represent the mean and standard deviation over three independent training.}
\label{tab:performance}
\end{table*}

To analyze the performance of OR-R1, we conduct experiments based on the open-source LLM Qwen3-8B and compare it with various learning-based methods on extensive datasets. The experiments aim to answer three questions:
\begin{itemize}
    \item \textbf{RQ1:} How does OR-R1 perform compared to existing learning-based methods across diverse real-world operations research benchmarks?
    \item \textbf{RQ2:} What is the impact of the TGRPO algorithm on model performance?
    \item \textbf{RQ3:} How do different reward formulations affect the performance of OR-R1?
\end{itemize}

\subsection{Experimental Setup}
\subsubsection{Datasets.}

To thoroughly evaluate the capabilities of OR-R1, we utilize a diverse set of optimization-related benchmarks that are collected from previously published sources. These datasets are chosen because they present unique challenges and comprehensively cover various aspects of optimization problem solving. The specific characteristics and sizes of each key test set are based on the operations and implementations referenced from the LLMOPT. These benchmarks collectively provide a comprehensive evaluation of LLMs in optimization modeling and solving. \textbf{NL4Opt} \cite{ramamonjison2023nl4opt} offers 230 linear programming word problems, including an ‘objective’ domain for generalization assessment. \textbf{Mamo} \cite{huang2024mamo} challenges LLMs with 652 \textbf{Easy LP} and 211 \textbf{Complex LP} instances requiring deeper mathematical reasoning. \textbf{NLP4LP} \cite{ahmaditeshnizi2024optimus} features 242 richly annotated linear and mixed-integer linear programming problems, mitigating data leakage. \textbf{ComplexOR} \cite{xiao2023chain} presents 18 complex real-world operation research problems with implicit constraints and domain-specific knowledge requirements. \textbf{IndustryOR} \cite{huang2025orlm} assesses performance on 100 real-world operation research problems. \textbf{OptiBench} \cite{yang2024optibench} includes 605 optimization modeling word problems across linear, non-linear, and tabular data types, evaluating iterative optimization. Finally, the \textbf{ICML Competition} \cite{yang2024benchmarking} track focuses on automated optimization problem solving, using 410 evaluable data points from its public leaderboard.

\subsubsection{Training DataSets.}

The training of OR-R1 primarily involves two critical phases: Supervised Fine-Tuning (SFT) and a TGRPO-based reinforcement learning stage. 

\begin{itemize}
\item \textbf{SFT stage:} We utilize \textbf{ORInstruct}, a public dataset from ORLM. It contains 3,000 synthetic samples, which corresponds to 1/10 of the full dataset used in ORLM.
\item \textbf{TGRPO stage:} We used unlabeled test data for training.
\end{itemize}

\subsubsection{Base Model.}
Our base model is Qwen3-8B \cite{yang2025qwen3}. It's chosen for its robust general-purpose language understanding and generation capabilities, demonstrating strong performance in both Math and Coding Tasks. Furthermore, its model size is comparable to that of previous related work, making it a suitable choice for our scenario.

\subsubsection{Baselines.}
We evaluate our method against several strong baselines, including both general LLMs and specialized optimization-focused models. Our primary baseline is the Qwen3-8B model fine-tuned on the ORInstruct(3K) dataset. Also, we compare with the following methods:

\begin{itemize}
\item \textbf{ORLM \cite{huang2025orlm}:}
This framework focuses on training open-source Large Language Models (LLMs) for optimization modeling and solver code development. It uses a semi-automated data synthesis framework called ORInstruct to generate high-quality training data from seed industry cases. The authors demonstrate results using Llama3-8B \cite{dubey2024llama} as their base model.

\item \textbf{LLMOPT \cite{jiang2024llmopt}:}
This framework employs multi-instruction tuning to improve problem formalization and solver code generation accuracy. It uses a ‘five-element formulation’ to define optimization problems and leverages data augmentation with expert / GPT4-based data labeling \cite{achiam2023gpt}. The framework incorporates supervised fine-tuning, model alignment, and a self-correction mechanism, with their primary results based on Qwen2.5-14B \cite{qwen2025qwen25technicalreport}.

\item \textbf{Base Model Variants:}
To ensure comprehensive comparison across different model architectures and sizes, we also include SFT-tuned versions of Qwen3-8B, Qwen2.5-7B \cite{qwen2025qwen25technicalreport} and Llama3-8B \cite{dubey2024llama} in our baseline evaluation. This helps isolate the impact of our methodology from the inherent capabilities of different base models.
\end{itemize}

\subsubsection{Evaluation Metrics.}
Aligned with ORLM, we adopt \textbf{Solution Accuracy} as our primary evaluation metric. For each optimization problem, the evaluation follows an end-to-end process: the LLM generates a response in natural language, which contains code blocks marked by \texttt{```}python ... \texttt{```}. A script automatically extracts and executes the generated Python code to obtain the predicted optimal objective value \(y_i\) for problem \(i\). Let \(y_i^*\) denote the ground truth optimal value. We define \textbf{Solution Accuracy} as:

\[
\text{Solution Accuracy} = \frac{1}{N} \sum_{i=1}^{N} \mathbb{I}(y_i = y_i^*)
\]

where \(N\) is the total number of problems, and \(\mathbb{I}(\cdot)\) is the indicator function, which equals 1 if the condition holds and 0 otherwise. In other words, a problem is considered correctly solved if and only if the predicted optimal value exactly matches the ground truth optimal value. This metric directly reflects the model’s end-to-end capability for generating and solving optimization problems.

\subsubsection{Implementation Details.}
The training process consists of two stages: SFT and TGRPO. In the SFT stage, Qwen3-8B is fine-tuned using the AdamW optimizer, a warmup-decay scheduler, and standard settings. In the TGRPO stage, the SFT output is further optimized using AdamW with a cosine scheduler and PEFT (LoRA). Both stages are trained on 4×A100 (40G) GPUs with BF16 precision. For all hyperparameters and detailed settings, please refer to the Appendix.

\subsection{Main Results}

Based on the experimental results (Table \ref{tab:performance}), our method demonstrates significant advantages across different test sets. Specifically, OR-R1 SFT(3K)-TGRPO achieves an average accuracy of 67.7\%, substantially outperforming all baseline models and achieving optimal performance on multiple test sets including NL4OPT, MAMO ComplexLP, IndustryOR, NLP4LP, OptiBench, and ICML Competition. Notably, even with only 100 samples for SFT, OR-R1 SFT(100)-TGRPO still achieves an average accuracy of 65.9\%, surpassing other baseline methods including ORLM and LLMOPT, which highlights the effectiveness of the TGRPO method. By comparing different base models, we find that Qwen3-8B, after SFT with 3K data, shows superior performance compared to Qwen2.5-7B and Llama3-8B. While this validates the importance of advanced base models, our TGRPO method can further improve performance by 3.1\%-6.4\%, demonstrating its effectiveness in optimizing operations research modeling and solving tasks.

\subsection{Training Dynamics of TGRPO}

\begin{figure}[h!]
\centering
\includegraphics[scale=0.34]{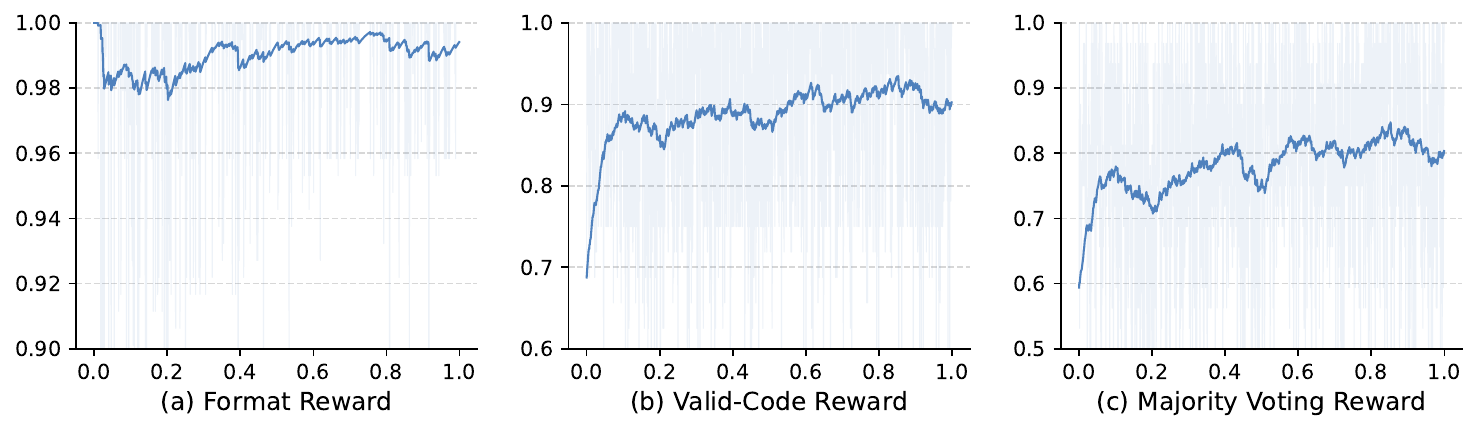} 
\caption{Overview of training dynamics for OR-R1 core reward components of SFT(3K)-TGRPO.}
\label{fig2}
\end{figure}

\begin{table*}[h!]
\renewcommand{\arraystretch}{1.5}
\centering
\resizebox{\textwidth}{!}
{
\begin{tabular}{lccccccccc}
\hline
\rowcolor{blue!10}
\textbf{Model} & \textbf{NL4OPT} & \textbf{MAMO} & \textbf{MAMO} & \textbf{IndustryOR} & \textbf{NLP4LP} & \textbf{ComplexOR} & \textbf{OptiBench} &\textbf{ICML} & \textbf{AVG}\\
\rowcolor{blue!10}
& & \textbf{EasyLP} & \textbf{ComplexLP} & & & & & \textbf{Competition} &\\
\hline
\multicolumn{10}{c}{\textit{Base Model Variants}} \\
\hline
\rowcolor{gray!20} 
Qwen3-8B SFT(3K) & 86.9 & 87.2 & 42.3 & 34.0 & 83.4 & 44.4 & 62.1 & 87.6 & 66.0 \\

\hline
\multicolumn{10}{c}{\textit{Ablation Studies (Base Model Qwen3-8B SFT(3K))}} \\
\hline

\rowcolor{gray!20} 

$+RL(R_{format})$
& $86.5_{\textcolor{blue}{\downarrow 0.4}}$ 
& $87.1_{\textcolor{blue}{\downarrow 0.1}}$ 
& $46.9_{\textcolor{red}{\uparrow 4.6}}$ 
& $34.0_{\textcolor{blue}{\downarrow 0.0}}$ 
& $83.5_{\textcolor{red}{\uparrow 0.1}}$ 
& $44.4_{\textcolor{blue}{\downarrow 0.0}}$ 
& $62.6_{\textcolor{red}{\uparrow 0.5}}$ 
& $87.3_{\textcolor{blue}{\downarrow 0.3}}$ 
& $66.5_{\textcolor{red}{\uparrow 0.5}}$ \\

$+RL(R_{code})$
& $86.5_{\textcolor{blue}{\downarrow 0.4}}$ 
& $\textbf{87.4}_{\textcolor{red}{\uparrow 0.2}}$ 
& $52.6_{\textcolor{red}{\uparrow 10.3}}$ 
& $\underline{38.0}_{\textcolor{red}{\uparrow 4.0}}$ 
& $84.3_{\textcolor{red}{\uparrow 0.8}}$ 
& $38.9_{\textcolor{blue}{\downarrow 5.5}}$ 
& $\underline{63.8}_{\textcolor{red}{\uparrow 1.7}}$ 
& $86.3_{\textcolor{blue}{\downarrow 1.3}}$ 
& $67.2_{\textcolor{red}{\uparrow 1.2}}$ \\

\rowcolor{gray!20} 
$+RL(R_{voting})$ 
& $\textbf{89.0}_{\textcolor{red}{\uparrow 2.1}}$ 
& $87.0_{\textcolor{blue}{\downarrow 0.2}}$ 
& $58.3_{\textcolor{red}{\uparrow 16.0}}$ 
& $33.0_{\textcolor{blue}{\downarrow 1.0}}$ 
& $84.7_{\textcolor{red}{\uparrow 1.3}}$ 
& $38.9_{\textcolor{blue}{\downarrow 5.5}}$ 
& $62.8_{\textcolor{red}{\uparrow 0.7}}$ 
& $\textbf{90.5}_{\textcolor{red}{\uparrow 2.9}}$ 
& $68.0_{\textcolor{red}{\uparrow 2.0}}$ \\

$+RL(R_{format} + R_{code})$ 
& $86.5_{\textcolor{blue}{\downarrow 0.4}}$ 
& $86.8_{\textcolor{blue}{\downarrow 0.4}}$ 
& $55.0_{\textcolor{red}{\uparrow 12.7}}$ 
& $35.0_{\textcolor{red}{\uparrow 1.0}}$ 
& $84.7_{\textcolor{red}{\uparrow 1.3}}$ 
& $44.4_{\textcolor{blue}{\downarrow 0.0}}$ 
& $63.1_{\textcolor{red}{\uparrow 1.0}}$ 
& $86.6_{\textcolor{blue}{\downarrow 1.0}}$ 
& $67.8_{\textcolor{red}{\uparrow 1.8}}$ \\

\rowcolor{gray!20} 
$+RL(R_{format} + R_{voting})$ 
& $87.8_{\textcolor{red}{\uparrow 0.9}}$ 
& $\underline{87.3}_{\textcolor{red}{\uparrow 0.1}}$ 
& $51.7_{\textcolor{red}{\uparrow 9.4}}$ 
& $33.0_{\textcolor{blue}{\downarrow 1.0}}$ 
& $\textbf{86.0}_{\textcolor{red}{\uparrow 2.6}}$ 
& $\textbf{50.0}_{\textcolor{red}{\uparrow 5.6}}$ 
& $\textbf{64.0}_{\textcolor{red}{\uparrow 1.9}}$ 
& $88.5_{\textcolor{red}{\uparrow 0.9}}$ 
& $68.5_{\textcolor{red}{\uparrow 2.5}}$ \\

$+RL(R_{code} + R_{voting})$ 
& $87.3_{\textcolor{red}{\uparrow 0.4}}$ 
& $87.1_{\textcolor{blue}{\downarrow 0.1}}$ 
& $\underline{67.8}_{\textcolor{red}{\uparrow 25.5}}$ 
& $\textbf{41.0}_{\textcolor{red}{\uparrow 7.0}}$ 
& $\underline{85.5}_{\textcolor{red}{\uparrow 2.1}}$ 
& $38.9_{\textcolor{blue}{\downarrow 5.5}}$ 
& $63.3_{\textcolor{red}{\uparrow 1.2}}$ 
& $87.6_{\textcolor{blue}{\downarrow 0.0}}$ 
& $\underline{69.7}_{\textcolor{red}{\uparrow 3.7}}$ \\

\rowcolor{gray!20} 
$+RL(R_{format} + R_{code} + R_{voting})$ 
& $\underline{88.6}_{\textcolor{red}{\uparrow 1.7}}$ 
& $87.0_{\textcolor{blue}{\downarrow 0.2}}$ 
& $\textbf{66.8}_{\textcolor{red}{\uparrow 24.5}}$ 
& $37.0_{\textcolor{red}{\uparrow 3.0}}$ 
& $84.3_{\textcolor{red}{\uparrow 0.9}}$ 
& $\textbf{50.0}_{\textcolor{red}{\uparrow 5.6}}$ 
& $63.5_{\textcolor{red}{\uparrow 1.4}}$ 
& $\underline{88.8}_{\textcolor{red}{\uparrow 1.2}}$ 
& $\textbf{70.8}_{\textcolor{red}{\uparrow 4.8}}$ \\

\hline
\end{tabular}
}
\caption{Ablation study on the reward formulations of OR-R1, using Qwen3-8B SFT(3K) as the base model. We report solution accuracy (\%) on eight operations research benchmarks and the overall average (AVG). Each row shows the effect of adding different reward components either individually or in combination. Performance gains or drops relative to the SFT-only baseline are marked in red (better) and blue (worse).
}
\label{tab:ablationstudies}
\end{table*}

Figure \ref{fig2} illustrates the training dynamics of the three core reward components (format, valid-code, and majority voting) for SFT(3K)-TGRPO. \textbf{Format Reward:} The format reward remains consistently high (above 0.98) throughout training, indicating that the model quickly learns to generate outputs in the correct format and maintains this ability stably.
\textbf{Valid-Code Reward:} The valid-code reward shows a clear upward trend in the early stages and gradually stabilizes around 0.9. This suggests that the model becomes increasingly capable of producing syntactically valid code as training progresses. \textbf{Majority Voting Reward:} The majority voting reward starts lower (around 0.7) but steadily improves, stabilizing near 0.8. This reflects the model's enhanced ability to generate solutions that are favored by majority voting, i.e., more frequently correct or consensus answers.
Across all three plots, the shaded regions indicate significant variance in reward values during training, but the overall trajectories of the moving averages are positive and stable. This demonstrates that SFT(3K)-TGRPO effectively optimizes all three reward components, leading to better structured, more valid, and more reliable model outputs.

\begin{figure}[h!]
\centering
\includegraphics[scale=0.35]{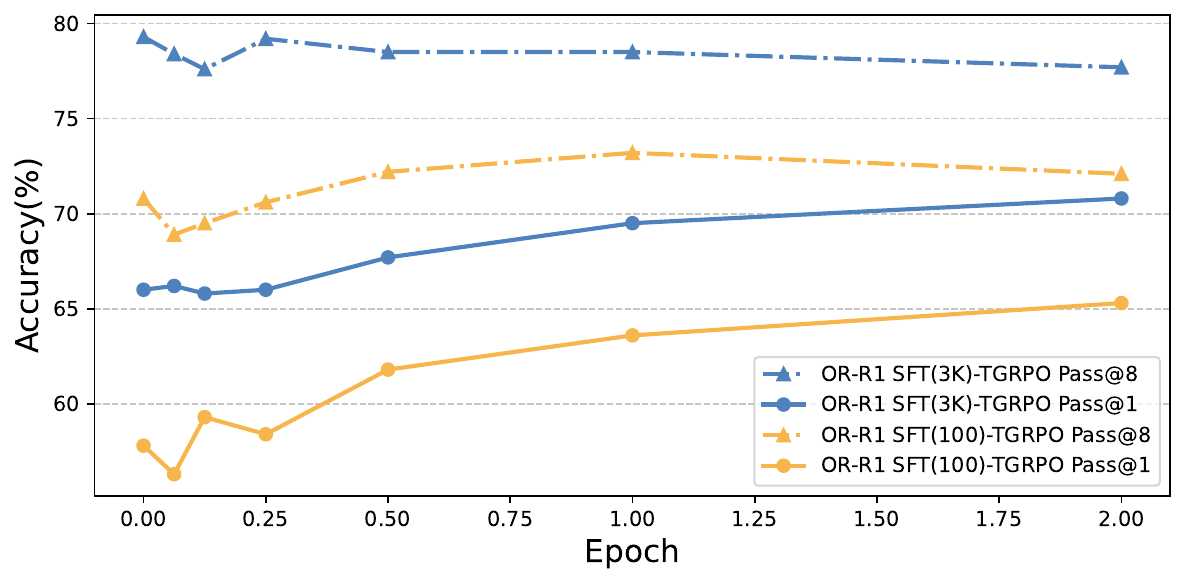}
\caption{Performance of Pass@1 and Pass@8 during TGRPO Training. Pass@1 measures the accuracy when only the model's top prediction is considered, while Pass@8 reflects the probability that at least one out of the top 8 generated solutions is correct.}
\label{fig3}
\end{figure}

Our motivation for developing TGRPO stems from the observed significant gap (13\%) between multiple sampling attempt(Pass@8) and single attempt(Pass@1) performance. While Pass@8 achieves impressive accuracy, Pass@1 initially performs substantially lower. This discrepancy indicates that the model possesses the underlying capability but lacks consistency in single-attempt scenarios. TGRPO was specifically designed to address this challenge by improving the model's deterministic performance and enhancing output consistency. As demonstrated in Fig. \ref{fig3}, TGRPO shows promising results in achieving this goal. After training, we successfully reduced this gap to 7\%. Pass@1 shows consistent improvement throughout the training process, indicating the effectiveness of our approach in enhancing single-generation reliability. It should be noted that the performance curve maintains an upward trajectory even at the end of our training iterations. Due to computational resource constraints, we had to limit the training duration. However, the steady positive trend suggests potential for improvements with extended training, as the model has not yet reached a performance plateau.

\subsection{Data Scale Effect on TGRPO}
\begin{figure}[h!]
\centering
\includegraphics[scale=0.42]{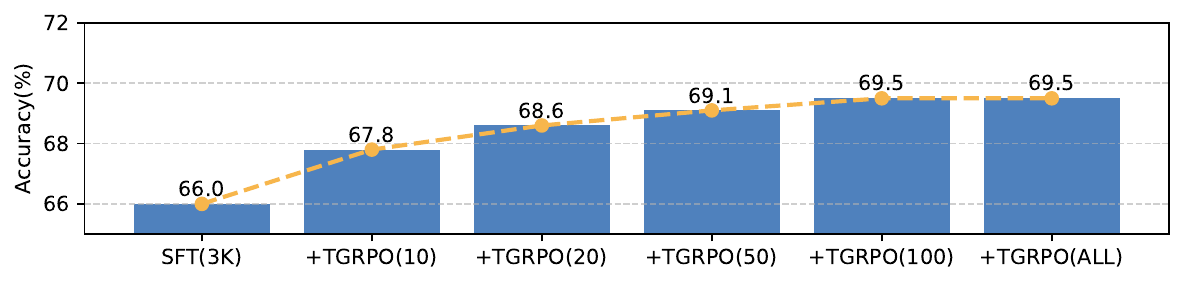}
\caption{The impact of different data scales on TGRPO performance. In TGRPO(N), N denotes the number of samples randomly selected from each test set for TGRPO training. Notably, all models here were trained for 160 steps.}
\label{fig4}
\end{figure}

Figure \ref{fig4} illustrates how model accuracy changes with different data scales for TGRPO training. As the number of TGRPO training samples (N) increases from 10 to 50, accuracy steadily improves from 66.0\% to 69.1\%. However, further increasing the data scale to include all available samples does not lead to higher accuracy. This suggests that TGRPO achieves significant performance gains with a relatively small amount of data, and increasing the data size further brings diminishing returns. In summary, TGRPO not only eliminates the need for additional labeled data, but also achieves strong training performance with only a small amount of in-domain data, showing its high data efficiency.

\subsection{Ablation Studies}

Table \ref{tab:ablationstudies} summarizes the impact of different reward components in OR-R1 using Qwen3-8B SFT(3K) as the base model. Adding individual rewards, such as format, code, or voting, each brings some improvement over the baseline of SFT only, with the voting reward showing the largest gain of a single component. Combining rewards further boosts performance: the best results are achieved when all three components are used together, yielding an average accuracy of 70.8\% (+4.8\% over baseline). These findings show that the rewards are complementary and that a comprehensive reward design is key to maximizing model performance.

\section{Conclusion}
This study introduces OR-R1, a data-efficient training framework for solving Operations Research (OR) optimization problems. By integrating Test-Time Group Relative Policy Optimization (TGRPO), OR-R1 achieves state-of-the-art performance, with an average accuracy of 67.7\% across multiple benchmarks, surpassing established methods like ORLM and LLMOPT. Remarkably, OR-R1 reaches competitive performance using only 100 labeled samples in the SFT stage, demonstrating its ability to drastically reduce data requirements while maintaining high accuracy. By leveraging tailored rewards, OR-R1 improves single-attempt reliability (Pass@1) and narrows the gap to multi-attempt performance (Pass@8) from 13\% to 7\%, ensuring consistent and robust outputs. This framework offers a scalable and cost-effective solution for training domain-specific large language models, which could benefit automated optimization applications in real-world industrial scenarios.

\bibliography{aaai2026}

@inproceedings{ramamonjison2023nl4opt,
  title={Nl4opt competition: Formulating optimization problems based on their natural language descriptions},
  author={Ramamonjison, Rindranirina and Yu, Timothy and Li, Raymond and Li, Haley and Carenini, Giuseppe and Ghaddar, Bissan and He, Shiqi and Mostajabdaveh, Mahdi and Banitalebi-Dehkordi, Amin and Zhou, Zirui and others},
  booktitle={NeurIPS 2022 Competition Track},
  pages={189--203},
  year={2023},
  organization={PMLR}
}

@inproceedings{xiao2023chain,
  title={Chain-of-experts: When llms meet complex operations research problems},
  author={Xiao, Ziyang and Zhang, Dongxiang and Wu, Yangjun and Xu, Lilin and Wang, Yuan Jessica and Han, Xiongwei and Fu, Xiaojin and Zhong, Tao and Zeng, Jia and Song, Mingli and others},
  booktitle={The twelfth international conference on learning representations},
  year={2023}
}

@article{ahmaditeshnizi2024optimus,
  title={Optimus: Scalable optimization modeling with (mi) lp solvers and large language models},
  author={AhmadiTeshnizi, Ali and Gao, Wenzhi and Udell, Madeleine},
  journal={arXiv preprint arXiv:2402.10172},
  year={2024}
}

@article{huang2024mamo,
  title={Mamo: a mathematical modeling benchmark with solvers},
  author={Huang, Xuhan and Shen, Qingning and Hu, Yan and Gao, Anningzhe and Wang, Benyou},
  journal={arXiv preprint arXiv:2405.13144},
  year={2024}
}

@article{huang2025orlm,
  title={Orlm: A customizable framework in training large models for automated optimization modeling},
  author={Huang, Chenyu and Tang, Zhengyang and Hu, Shixi and Jiang, Ruoqing and Zheng, Xin and Ge, Dongdong and Wang, Benyou and Wang, Zizhuo},
  journal={Operations Research},
  year={2025},
  publisher={INFORMS}
}

@article{yang2024optibench,
  title={OptiBench meets ReSocratic: Measure and improve LLMs for optimization modeling},
  author={Yang, Zhicheng and Wang, Yiwei and Huang, Yinya and Guo, Zhijiang and Shi, Wei and Han, Xiongwei and Feng, Liang and Song, Linqi and Liang, Xiaodan and Tang, Jing},
  journal={arXiv preprint arXiv:2407.09887},
  year={2024}
}

@article{jiang2024llmopt,
  title={LLMOPT: Learning to Define and Solve General Optimization Problems from Scratch},
  author={Jiang, Caigao and Shu, Xiang and Qian, Hong and Lu, Xingyu and Zhou, Jun and Zhou, Aimin and Yu, Yang},
  journal={arXiv preprint arXiv:2410.13213},
  year={2024}
}

@inproceedings{mostajabdaveh2025evaluating,
  title={Evaluating LLM Reasoning in the Operations Research Domain with ORQA},
  author={Mostajabdaveh, Mahdi and Yu, Timothy Tin Long and Dash, Samarendra Chandan Bindu and Ramamonjison, Rindra and Byusa, Jabo Serge and Carenini, Giuseppe and Zhou, Zirui and Zhang, Yong},
  booktitle={Proceedings of the AAAI Conference on Artificial Intelligence},
  volume={39},
  number={23},
  pages={24902--24910},
  year={2025}
}

@article{zhang2025or,
  title={Or-llm-agent: Automating modeling and solving of operations research optimization problem with reasoning large language model},
  author={Zhang, Bowen and Luo, Pengcheng},
  journal={arXiv preprint arXiv:2503.10009},
  year={2025}
}

@article{thind2025optimai,
  title={OptimAI: Optimization from Natural Language Using LLM-Powered AI Agents},
  author={Thind, Raghav and Sun, Youran and Liang, Ling and Yang, Haizhao},
  journal={arXiv preprint arXiv:2504.16918},
  year={2025}
}

@article{yang2025qwen3,
  title={Qwen3 technical report},
  author={Yang, An and Li, Anfeng and Yang, Baosong and Zhang, Beichen and Hui, Binyuan and Zheng, Bo and Yu, Bowen and Gao, Chang and Huang, Chengen and Lv, Chenxu and others},
  journal={arXiv preprint arXiv:2505.09388},
  year={2025}
}

@article{shao2024deepseekmath,
  title={Deepseekmath: Pushing the limits of mathematical reasoning in open language models},
  author={Shao, Zhihong and Wang, Peiyi and Zhu, Qihao and Xu, Runxin and Song, Junxiao and Bi, Xiao and Zhang, Haowei and Zhang, Mingchuan and Li, YK and Wu, Y and others},
  journal={arXiv preprint arXiv:2402.03300},
  year={2024}
}

@article{zuo2025ttrl,
  title={Ttrl: Test-time reinforcement learning},
  author={Zuo, Yuxin and Zhang, Kaiyan and Sheng, Li and Qu, Shang and Cui, Ganqu and Zhu, Xuekai and Li, Haozhan and Zhang, Yuchen and Long, Xinwei and Hua, Ermo and others},
  journal={arXiv preprint arXiv:2504.16084},
  year={2025}
}

@article{prabhudesai2025maximizing,
  title={Maximizing Confidence Alone Improves Reasoning},
  author={Prabhudesai, Mihir and Chen, Lili and Ippoliti, Alex and Fragkiadaki, Katerina and Liu, Hao and Pathak, Deepak},
  journal={arXiv preprint arXiv:2505.22660},
  year={2025}
}

@article{schulman2017proximal,
  title={Proximal policy optimization algorithms},
  author={Schulman, John and Wolski, Filip and Dhariwal, Prafulla and Radford, Alec and Klimov, Oleg},
  journal={arXiv preprint arXiv:1707.06347},
  year={2017}
}

@article{yang2024benchmarking,
  title={Benchmarking LLMs for Optimization Modeling and Enhancing Reasoning via Reverse Socratic Synthesis},
  author={Yang, Zhicheng and Huang, Yinya and Shi, Wei and Feng, Liang and Song, Linqi and Wang, Yiwei and Liang, Xiaodan and Tang, Jing},
  journal={arXiv preprint arXiv:2407.09887},
  year={2024}
}

@article{lu2025optmath,
  title={Optmath: A scalable bidirectional data synthesis framework for optimization modeling},
  author={Lu, Hongliang and Xie, Zhonglin and Wu, Yaoyu and Ren, Can and Chen, Yuxuan and Wen, Zaiwen},
  journal={arXiv preprint arXiv:2502.11102},
  year={2025}
}

@misc{qwen2025qwen25technicalreport,
      title={Qwen2.5 Technical Report}, 
      author={Qwen and : and An Yang and Baosong Yang and Beichen Zhang and Binyuan Hui and Bo Zheng and Bowen Yu and Chengyuan Li and Dayiheng Liu and Fei Huang and Haoran Wei and Huan Lin and Jian Yang and Jianhong Tu and Jianwei Zhang and Jianxin Yang and Jiaxi Yang and Jingren Zhou and Junyang Lin and Kai Dang and Keming Lu and Keqin Bao and Kexin Yang and Le Yu and Mei Li and Mingfeng Xue and Pei Zhang and Qin Zhu and Rui Men and Runji Lin and Tianhao Li and Tianyi Tang and Tingyu Xia and Xingzhang Ren and Xuancheng Ren and Yang Fan and Yang Su and Yichang Zhang and Yu Wan and Yuqiong Liu and Zeyu Cui and Zhenru Zhang and Zihan Qiu},
      year={2025},
      eprint={2412.15115},
      archivePrefix={arXiv},
      primaryClass={cs.CL},
      url={https://arxiv.org/abs/2412.15115}, 
}

@article{dubey2024llama,
  title={The llama 3 herd of models},
  author={Dubey, Abhimanyu and Jauhri, Abhinav and Pandey, Abhinav and Kadian, Abhishek and Al-Dahle, Ahmad and Letman, Aiesha and Mathur, Akhil and Schelten, Alan and Yang, Amy and Fan, Angela and others},
  journal={arXiv e-prints},
  pages={arXiv--2407},
  year={2024}
}

@article{cobbe2021training,
  title={Training verifiers to solve math word problems},
  author={Cobbe, Karl and Kosaraju, Vineet and Bavarian, Mohammad and Chen, Mark and Jun, Heewoo and Kaiser, Lukasz and Plappert, Matthias and Tworek, Jerry and Hilton, Jacob and Nakano, Reiichiro and others},
  journal={arXiv preprint arXiv:2110.14168},
  year={2021}
}

@article{hendrycks2021measuring,
  title={Measuring mathematical problem solving with the math dataset},
  author={Hendrycks, Dan and Burns, Collin and Kadavath, Saurav and Arora, Akul and Basart, Steven and Tang, Eric and Song, Dawn and Steinhardt, Jacob},
  journal={arXiv preprint arXiv:2103.03874},
  year={2021}
}

@article{he2024olympiadbench,
  title={Olympiadbench: A challenging benchmark for promoting agi with olympiad-level bilingual multimodal scientific problems},
  author={He, Chaoqun and Luo, Renjie and Bai, Yuzhuo and Hu, Shengding and Thai, Zhen Leng and Shen, Junhao and Hu, Jinyi and Han, Xu and Huang, Yujie and Zhang, Yuxiang and others},
  journal={arXiv preprint arXiv:2402.14008},
  year={2024}
}

@article{li2022competition,
  title={Competition-Level Code Generation with AlphaCode},
    author={Li, Yujia and Choi, David and Chung, Junyoung and Kushman, Nate and
    Schrittwieser, Julian and Leblond, R{\'e}mi and Eccles, Tom and
    Keeling, James and Gimeno, Felix and Dal Lago, Agustin and
    Hubert, Thomas and Choy, Peter and de Masson d'Autume, Cyprien and
    Babuschkin, Igor and Chen, Xinyun and Huang, Po-Sen and Welbl, Johannes and
    Gowal, Sven and Cherepanov, Alexey and Molloy, James and
    Mankowitz, Daniel and Sutherland Robson, Esme and Kohli, Pushmeet and
    de Freitas, Nando and Kavukcuoglu, Koray and Vinyals, Oriol},
  journal={arXiv preprint arXiv:2203.07814},
  year={2022}
}

@misc{codealpaca,
  author = {Sahil Chaudhary},
  title = {Code Alpaca: An Instruction-following LLaMA model for code generation},
  year = {2023},
  publisher = {GitHub},
  journal = {GitHub repository},
  howpublished = {\url{https://github.com/sahil280114/codealpaca}},
}

@article{brown2020language,
  title={Language models are few-shot learners},
  author={Brown, Tom and Mann, Benjamin and Ryder, Nick and Subbiah, Melanie and Kaplan, Jared D and Dhariwal, Prafulla and Neelakantan, Arvind and Shyam, Pranav and Sastry, Girish and Askell, Amanda and others},
  journal={Advances in neural information processing systems},
  volume={33},
  pages={1877--1901},
  year={2020}
}

@article{wei2022chain,
  title={Chain-of-thought prompting elicits reasoning in large language models},
  author={Wei, Jason and Wang, Xuezhi and Schuurmans, Dale and Bosma, Maarten and Xia, Fei and Chi, Ed and Le, Quoc V and Zhou, Denny and others},
  journal={Advances in neural information processing systems},
  volume={35},
  pages={24824--24837},
  year={2022}
}

@article{yao2023tree,
  title={Tree of thoughts: Deliberate problem solving with large language models},
  author={Yao, Shunyu and Yu, Dian and Zhao, Jeffrey and Shafran, Izhak and Griffiths, Tom and Cao, Yuan and Narasimhan, Karthik},
  journal={Advances in neural information processing systems},
  volume={36},
  pages={11809--11822},
  year={2023}
}

@inproceedings{besta2024graph,
  title={Graph of thoughts: Solving elaborate problems with large language models},
  author={Besta, Maciej and Blach, Nils and Kubicek, Ales and Gerstenberger, Robert and Podstawski, Michal and Gianinazzi, Lukas and Gajda, Joanna and Lehmann, Tomasz and Niewiadomski, Hubert and Nyczyk, Piotr and others},
  booktitle={Proceedings of the AAAI Conference on Artificial Intelligence},
  volume={38},
  number={16},
  pages={17682--17690},
  year={2024}
}

@article{ouyang2022training,
  title={Training language models to follow instructions with human feedback},
  author={Ouyang, Long and Wu, Jeffrey and Jiang, Xu and Almeida, Diogo and Wainwright, Carroll and Mishkin, Pamela and Zhang, Chong and Agarwal, Sandhini and Slama, Katarina and Ray, Alex and others},
  journal={Advances in neural information processing systems},
  volume={35},
  pages={27730--27744},
  year={2022}
}

@article{christiano2017deep,
  title={Deep reinforcement learning from human preferences},
  author={Christiano, Paul F and Leike, Jan and Brown, Tom and Martic, Miljan and Legg, Shane and Amodei, Dario},
  journal={Advances in neural information processing systems},
  volume={30},
  year={2017}
}

@article{rafailov2023direct,
  title={Direct preference optimization: Your language model is secretly a reward model},
  author={Rafailov, Rafael and Sharma, Archit and Mitchell, Eric and Manning, Christopher D and Ermon, Stefano and Finn, Chelsea},
  journal={Advances in Neural Information Processing Systems},
  volume={36},
  pages={53728--53741},
  year={2023}
}

@article{guo2025deepseek,
  title={Deepseek-r1: Incentivizing reasoning capability in llms via reinforcement learning},
  author={Guo, Daya and Yang, Dejian and Zhang, Haowei and Song, Junxiao and Zhang, Ruoyu and Xu, Runxin and Zhu, Qihao and Ma, Shirong and Wang, Peiyi and Bi, Xiao and others},
  journal={arXiv preprint arXiv:2501.12948},
  year={2025}
}

@article{yu2025dapo,
  title={Dapo: An open-source llm reinforcement learning system at scale},
  author={Yu, Qiying and Zhang, Zheng and Zhu, Ruofei and Yuan, Yufeng and Zuo, Xiaochen and Yue, Yu and Fan, Tiantian and Liu, Gaohong and Liu, Lingjun and Liu, Xin and others},
  journal={arXiv preprint arXiv:2503.14476},
  year={2025}
}

@article{sheng2024hybridflow,
  title   = {HybridFlow: A Flexible and Efficient RLHF Framework},
  author  = {Guangming Sheng and Chi Zhang and Zilingfeng Ye and Xibin Wu and Wang Zhang and Ru Zhang and Yanghua Peng and Haibin Lin and Chuan Wu},
  year    = {2024},
  journal = {arXiv preprint arXiv: 2409.19256}
}

@inproceedings{zhang2024framework,
  title={A Framework for Training Large Language Models for Code Generation via Proximal Policy Optimization},
  author={Zhang, Chi and Sheng, Guangming and Liu, Siyao and Li, Jiahao and Feng, Ziyuan and Liu, Zherui and Liu, Xin and Jia, Xiaoying and Peng, Yanghua and Lin, Haibin and others},
  booktitle={NL2Code Workshop of ACM KDD},
  year={2024}
}

@article{kumar2024training,
  title={Training language models to self-correct via reinforcement learning},
  author={Kumar, Aviral and Zhuang, Vincent and Agarwal, Rishabh and Su, Yi and Co-Reyes, John D and Singh, Avi and Baumli, Kate and Iqbal, Shariq and Bishop, Colton and Roelofs, Rebecca and others},
  journal={arXiv preprint arXiv:2409.12917},
  year={2024}
}

@article{qu2024recursive,
  title={Recursive introspection: Teaching language model agents how to self-improve},
  author={Qu, Yuxiao and Zhang, Tianjun and Garg, Naman and Kumar, Aviral},
  journal={Advances in Neural Information Processing Systems},
  volume={37},
  pages={55249--55285},
  year={2024}
}

@article{xin2024deepseek,
  title={Deepseek-prover-v1. 5: Harnessing proof assistant feedback for reinforcement learning and monte-carlo tree search},
  author={Xin, Huajian and Ren, ZZ and Song, Junxiao and Shao, Zhihong and Zhao, Wanjia and Wang, Haocheng and Liu, Bo and Zhang, Liyue and Lu, Xuan and Du, Qiushi and others},
  journal={arXiv preprint arXiv:2408.08152},
  year={2024}
}

@article{wang2020tent,
  title={Tent: Fully test-time adaptation by entropy minimization},
  author={Wang, Dequan and Shelhamer, Evan and Liu, Shaoteng and Olshausen, Bruno and Darrell, Trevor},
  journal={arXiv preprint arXiv:2006.10726},
  year={2020}
}

@misc{jiang2024surveylargelanguagemodels,
      title={A Survey on Large Language Models for Code Generation}, 
      author={Juyong Jiang and Fan Wang and Jiasi Shen and Sungju Kim and Sunghun Kim},
      year={2024},
      eprint={2406.00515},
      archivePrefix={arXiv},
      primaryClass={cs.CL},
      url={https://arxiv.org/abs/2406.00515}, 
}

@misc{zhao2025surveylargelanguagemodels,
      title={A Survey of Large Language Models}, 
      author={Wayne Xin Zhao and Kun Zhou and Junyi Li and Tianyi Tang and Xiaolei Wang and Yupeng Hou and Yingqian Min and Beichen Zhang and Junjie Zhang and Zican Dong and Yifan Du and Chen Yang and Yushuo Chen and Zhipeng Chen and Jinhao Jiang and Ruiyang Ren and Yifan Li and Xinyu Tang and Zikang Liu and Peiyu Liu and Jian-Yun Nie and Ji-Rong Wen},
      year={2025},
      eprint={2303.18223},
      archivePrefix={arXiv},
      primaryClass={cs.CL},
      url={https://arxiv.org/abs/2303.18223}, 
}

@article{brucker1999resource,
  title={Resource-constrained project scheduling: Notation, classification, models, and methods},
  author={Brucker, Peter and Drexl, Andreas and M{\"o}hring, Rolf and Neumann, Klaus and Pesch, Erwin},
  journal={European journal of operational research},
  volume={112},
  number={1},
  pages={3--41},
  year={1999},
  publisher={Elsevier}
}

@misc{chen2023universalselfconsistencylargelanguage,
      title={Universal Self-Consistency for Large Language Model Generation}, 
      author={Xinyun Chen and Renat Aksitov and Uri Alon and Jie Ren and Kefan Xiao and Pengcheng Yin and Sushant Prakash and Charles Sutton and Xuezhi Wang and Denny Zhou},
      year={2023},
      eprint={2311.17311},
      archivePrefix={arXiv},
      primaryClass={cs.CL},
      url={https://arxiv.org/abs/2311.17311}, 
}

@misc{huang2024enhancinglargelanguagemodels,
      title={Enhancing Large Language Models in Coding Through Multi-Perspective Self-Consistency}, 
      author={Baizhou Huang and Shuai Lu and Weizhu Chen and Xiaojun Wan and Nan Duan},
      year={2024},
      eprint={2309.17272},
      archivePrefix={arXiv},
      primaryClass={cs.CL},
      url={https://arxiv.org/abs/2309.17272}, 
}

@article{hurst2024gpt,
  title={Gpt-4o system card},
  author={Hurst, Aaron and Lerer, Adam and Goucher, Adam P and Perelman, Adam and Ramesh, Aditya and Clark, Aidan and Ostrow, AJ and Welihinda, Akila and Hayes, Alan and Radford, Alec and others},
  journal={arXiv preprint arXiv:2410.21276},
  year={2024}
}

@misc{ahn2024largelanguagemodelsmathematical,
      title={Large Language Models for Mathematical Reasoning: Progresses and Challenges}, 
      author={Janice Ahn and Rishu Verma and Renze Lou and Di Liu and Rui Zhang and Wenpeng Yin},
      year={2024},
      eprint={2402.00157},
      archivePrefix={arXiv},
      primaryClass={cs.CL},
      url={https://arxiv.org/abs/2402.00157}, 
}

@article{bairi2024codeplan,
  title={Codeplan: Repository-level coding using llms and planning},
  author={Bairi, Ramakrishna and Sonwane, Atharv and Kanade, Aditya and C, Vageesh D and Iyer, Arun and Parthasarathy, Suresh and Rajamani, Sriram and Ashok, Balasubramanyan and Shet, Shashank},
  journal={Proceedings of the ACM on Software Engineering},
  volume={1},
  number={FSE},
  pages={675--698},
  year={2024},
  publisher={ACM New York, NY, USA}
}

@article{achiam2023gpt,
  title={Gpt-4 technical report},
  author={Achiam, Josh and Adler, Steven and Agarwal, Sandhini and Ahmad, Lama and Akkaya, Ilge and Aleman, Florencia Leoni and Almeida, Diogo and Altenschmidt, Janko and Altman, Sam and Anadkat, Shyamal and others},
  journal={arXiv preprint arXiv:2303.08774},
  year={2023}
}

@book{harrison2019logistics,
  title={Logistics management and strategy},
  author={Harrison, Alan and Van Hoek, Remko and Skipworth, Heather and Aitken, James},
  year={2019},
  publisher={Pearson UK}
}

@article{bretthauer1995nonlinear,
  title={The nonlinear resource allocation problem},
  author={Bretthauer, Kurt M and Shetty, Bala},
  journal={Operations research},
  volume={43},
  number={4},
  pages={670--683},
  year={1995},
  publisher={INFORMS}
}

@article{long2020dynamic,
  title={Dynamic scheduling of multiclass many-server queues with abandonment: The generalized c$\mu$/h rule},
  author={Long, Zhenghua and Shimkin, Nahum and Zhang, Hailun and Zhang, Jiheng},
  journal={Operations Research},
  volume={68},
  number={4},
  pages={1218--1230},
  year={2020},
  publisher={INFORMS}
}

@article{lee2015impact,
  title={The impact of slow ocean steaming on delivery reliability and fuel consumption},
  author={Lee, Chung-Yee and Lee, Hau L and Zhang, Jiheng},
  journal={Transportation Research Part E: Logistics and Transportation Review},
  volume={76},
  pages={176--190},
  year={2015},
  publisher={Elsevier}
}

\end{document}